# Agentic AI for operating scientific instruments for nanoscale characterization

Zahra Ayar[1], Marcos Penedo[1,*], Mahdi Mehdikhani[1], Nahid Hosseini[1], Prabhu Prasad Swain[1], Georg E. Fantner[1,*]

*Laboratory for Bio- and Nano-Instrumentation, Institute of Bioengineering, School of Engineering, Swiss Federal Institute of Technology Lausanne (EPFL), Lausanne 1015, Switzerland*

**Corresponding author. E-mail: georg.fantner@epfl.ch, marcos.penedo@epfl.ch*

**Abstract**

Operating a scientific instrument such as an atomic force microscope (AFM) requires continuous expert decision-making. A trained user defines the experimental intent, translates it into instrument commands, assesses incoming data, adjusts imaging parameters, and post-processes the final image. Existing automation usually addresses only parts of this workflow through hard-coded routines, task-specific controllers, or trained machine-learning models. Here we present an agentic-AI framework that operates the executable part of the AFM workflow using a general-purpose, tool-augmented large language model connected to instrument functions through the Model Context Protocol (MCP). The framework consists of three MCP-based agents: *AFM Messenger* converts natural-language instructions into checked instrument commands; *AFM Pilot* assesses image quality through a large language model (LLM) and, if necessary, adapts imaging parameters; and *AFM Doctor* diagnoses image artifacts and applies transparent post-processing from a pre-approved tool set. Because the language model performs image assessment rather than a fixed scalar objective or external optimizer, the same strategy can be applied across sample types and imaging modes without specific retraining. Safe hardware operation is enforced through an ambiguity check layer before execution. Benchmarking against fine-tuned and off-the-shelf tool-using models shows that this guarded execution layer, rather than model capability alone, reduces wrong-command execution to zero. In live experiments on different samples, *AFM Pilot* matched expert operators in image quality, iteration count, and tuning time, with no significant difference. These results demonstrate a safe route to agentic operation of scientific instruments, where experimental intent remains human-defined while command execution, image-based tuning, and post-processing are delegated to AI agents.

**Keywords:** *agentic AI, tool-augmented LLMs, Model Context Protocol (MCP), automated microscopy, atomic force microscopy (AFM), closed-loop parameter tuning, failure analysis, ambiguity handling, artifact diagnosis, human-in-the-loop control*

## 1. Introduction

Artificial intelligence is increasingly transforming how experiments are designed and executed, moving from tools that assist single steps toward agents that plan and carry out multi-step scientific work[1–5]. This marks a transition from AI for science toward agentic science, in which autonomous systems in chemistry[5–7], materials synthesis[5,8–10], and laboratory robotics[11–13] can coordinate instruments and interpret their own results. These examples point to a broader opportunity: using AI not only to analyze scientific data after acquisition, but to operate the physical instruments that generate those data.

Microscopy and imaging are natural areas for this transition because acquiring an image is not simply a passive recording step, but an active experimental process. Conventional machine-learning methods have been used in X-ray imaging[14], optical microcopy[15,16], and electron microscopy[17,18], to support automated acquisition, reconstruction, analysis, or experimental decision-making. In scanning probe microscopy, earlier approaches used trained machine learning models to guide the measurement area[19], tune imaging conditions[20], and apply methods to autonomous operation[21] and data analysis[22]. These studies show the potential of automation in microscopy, but most existing systems still address selected parts of the workflow. They typically depend on a task-specific model, reward function, or control routine, and do not fully address the complexity of imaging tasks.

Atomic force microscopy (AFM)[23] exemplifies both the need and the difficulty. Image quality depends on feedback and scan parameters that must be adjusted continuously as the tip, sample, and feedback response change between different experiments. Artifacts such as parachuting[24], feedback ringing[25], hysteresis[26,27], creep[26,27], double-tip[28], and loss of surface tracking require different corrective actions, with trade-offs between image quality, scan speed, and sample safety[29–31]. Although conventional automation and task-specific machine learning have addressed parts of this workflow, the image-driven assessment needed to connect command execution, parameter tuning, and post-processing has largely remained human. This makes AFM a natural test case for agentic instrument control.

Large language models (LLMs) have recently begun to be explored for microscopy operation and automated experiment design[32]. Two architectural strategies are particularly relevant. In the first, a model is fine-tuned[33–35] on instrument-specific data to generate instrument commands directly. This embeds instrument knowledge in the model weights, but requires preparing large prompt-response datasets and retraining whenever the instrument software or base model changes. It also offers limited protection against silent execution errors[13,32]. In the second, a general-purpose model is connected to predefined functions through a structured interface such as the Model Context Protocol (MCP)[36–38]. In this architecture, tool selection is explicit. At the same time, unit definitions, required parameters, and clarification rules can be enforced outside the model. The MCP approach needs only a description of the available tools and their parameters, which can be written in hours and reused across models without retraining.

The most comprehensive prior demonstration of LLM-operated AFM is that of Mandal et al.[39], whose artificially intelligent lab assistant (AILA) used off-the-shelf models with LangChain/LangGraph orchestration. In that system, the LLM interpreted tasks and experimental outputs, while a genetic algorithm optimized image quality by tuning proportional–integral–derivative feedback gains to maximize trace-retrace structural similarity. The study also raised important safety concerns, reporting about 65% of task success for the best model and showing that ambiguous instructions could lead to erratic or incorrect commands. Building on this foundation, we asked whether an LLM-based agent can move beyond workflow coordination to directly assess AFM image quality and select corrective actions from the acquired data.

In this work, we try to replace the human AFM expert in some aspects of the AFM imaging workflow with agentic AI. We present an MCP-based agentic AFM system that translates human natural-language commands into instrument actions, evaluates AFM images and associated experimental data using the multimodal vision-language capabilities of the LLM (hereafter, "*AI-vision*"), and adjusts imaging parameters without relying on a conventional control model or a fixed scalar optimization metric.

We implemented this framework for AFM through three cooperating agents: *AFM Messenger* for communicating with AFM hardware, *AFM Pilot* for closed-loop image parameter optimization, and *AFM Doctor* for artifact diagnosis and transparent post-processing. We benchmark the execution layer, test live agentic imaging, and compare its operation with expert practice. Because incorrect or ambiguous commands can damage the probe, sample, or instrument, the execution layer checks each request before acting and asks for clarification when the prompt is underspecified. We frame this deliberately as agentic operation of a scientific instrument, not a self-driving laboratory, where the experimental intent remains with the human, but the agentic system carries out command execution, imaging parameter tuning, and post-processing.

# 2. Results

## 2.1 An agentic framework for the scientific instrument operation

When a trained user operates an AFM, their actions can be grouped into four tasks: 1) deciding what to measure (intent), 2) translating that intent into instrument commands, 3) assessing the incoming data and tuning parameters until the image is "as good as it gets", and 4) post-processing the final data to extract the desired information. In conventional AFM operation, a human is responsible for all actions, from intent and prompts to interacting with the human-machine interface, image assessment, parameter optimization, and image post-processing (Figure 1a).

Here, we designed a framework for AFM that divides responsibilities between the human operator and the agentic system (Figure 1b). The human retains the experimental intent and decides what to measure, while the agentic system handles the executable parts of the operation. In this way, the framework targets instrument operation rather than experimental goal selection.

The architecture consists of three MCP-based agents (Figure 1b). The first MCP-based agent, "*AFM Messenger*", translates user instructions into safe instrument actions. The LLM interprets the prompt request and converts it into instrument commands using tools imported from the Python library that controls the hardware. The prompt is then screened for ambiguity before execution, since underspecified instructions can lead to erratic or potentially damaging actions on physical hardware. Ambiguous prompts are returned to the operator for clarification, while confirmed prompts proceed to execution. A representative of a session is provided in Supplementary Data - Information S2.

The second agent, "*AFM Pilot*", performs image-based parameter tuning. *AFM Pilot* operates in a loop, retrieving recent height and error data from the instrument, preparing diagnostic views, and using *AI vision* to assess whether the measurement shows signs of non-optimal imaging, such as parachuting or feedback ringing. These artifacts indicate insufficient surface tracking or reduced feedback stability. Based on this assessment, the agent evaluates the current imaging quality and proposes bounded parameter updates to improve surface tracking while maintaining stable feedback. The agent then executes the proposed parameter updates through *AFM Messenger*.

The third agent, "*AFM Doctor*", performs post-processing. *AFM Doctor* inspects the final image, identifies common artifacts, and selects corrections from a pre-approved set of transparent processing tools. By separating command execution, imaging parameter optimization, and post-processing into independent MCP-based agents, the framework keeps each layer testable. It constrains the actions available to the agent while allowing the same operating logic to be reused across imaging modes.

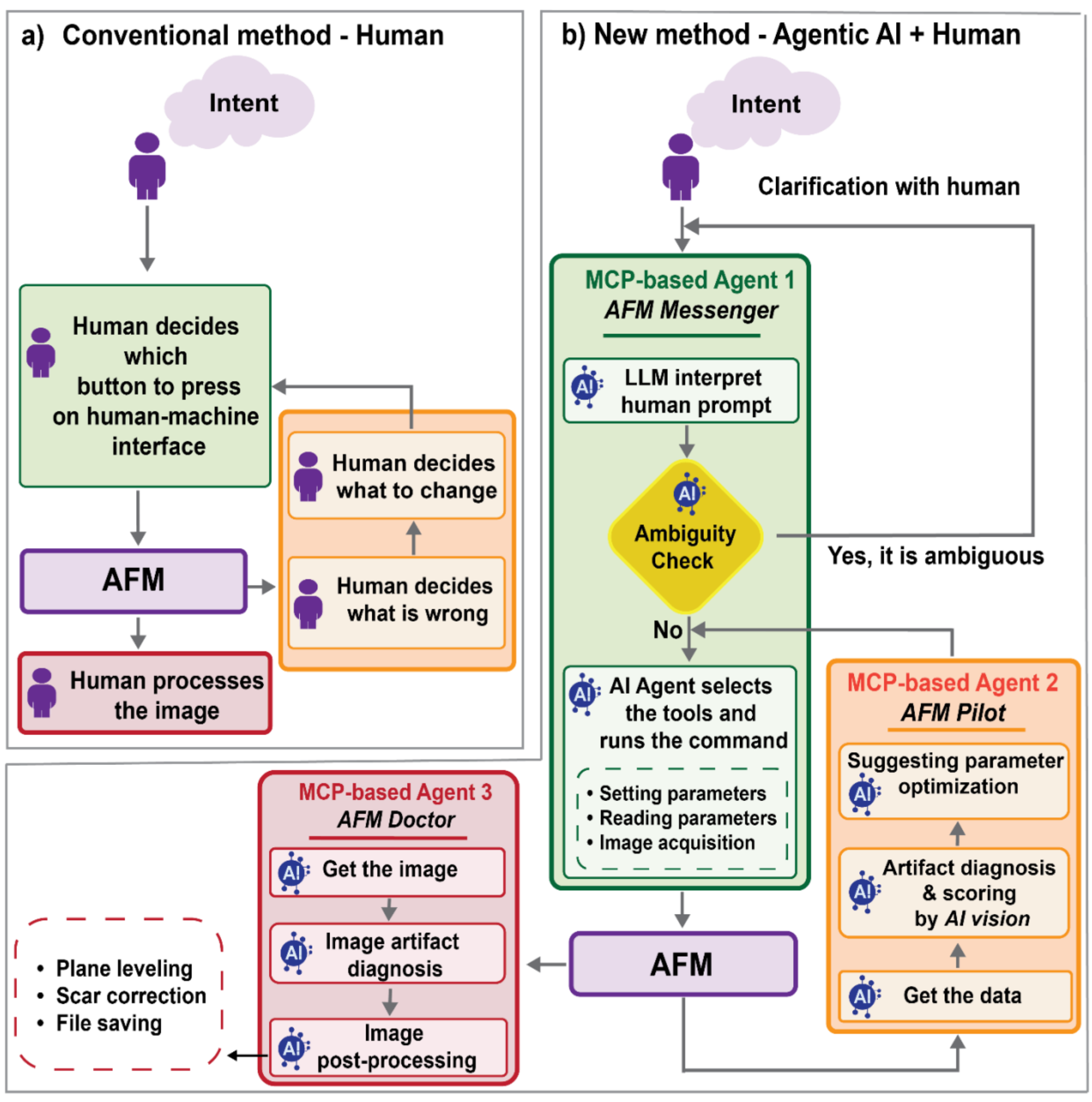

**Figure 1.** The conventional operator-in-the-loop pipeline vs the agentic-AI framework. a) In conventional operation, the human operates the control interface, evaluates the acquired data and image quality, and decides the appropriate next steps, including parameter adjustments and post-processing. b) In the agentic framework, the human retains the experimental intent, while MCP-based agents handle command execution, imaging parameter tuning, and post-processing. The system returns ambiguous commands to the operator for clarification before execution. Confirmed commands are executed through *AFM Messenger*, *AFM Pilot* uses *AI-vision* to assess AFM images and associated data and tune imaging parameters, and *AFM Doctor* applies pre-approved transparent post-processing steps. *AFM Doctor* applies pre-approved transparent post-processing steps.

## 2.2 Interactive ambiguity checking enables safe operation

Before allowing the agent to modify parameters on real hardware, we tested how LLM requests could be translated into executable AFM actions and which safeguards were required before execution (Figure 2). All configurations ultimately execute commands through RePySPM[40], our Python control library for the AFM, allowing us to compare different ways of connecting the language model to the same instrument interface.

We evaluated the configurations using the same benchmark of AFM commands, which an experienced AFM user manually designed and reviewed. The benchmark data included individual and consecutive commands, different instrument functions, unit and value variations, and under-specified requests. Details of the benchmark construction are provided in Methods Section 5.3, and the complete dataset is available in GitHub repository [41].

In the fine-tuned approach, we trained GPT-4.1 on instrument-specific command examples (Figure 2a). We evaluated direct command generation (FT-GPT) and the same fine-tuned model with function calling (FT-GPT-FC). In the second approach, the model was given structured access to 129 tools and selected the appropriate tool at runtime. We evaluated this using Claude Sonnet 4.6 and Claude Opus 4.8.

Without access to tools, the unmodified Claude Sonnet 4.6 model failed on 89.2% of commands (Figure 2b, Supplementary Data – Table S1). In contrast, fine-tuned GPT (FT-GPT) had an error rate of 28.4 ± 1.5%, which fell to 23.3 ± 0.7% with function calling. A direct comparison with a fine-tuned model was not possible because Claude no longer provides this feature. However, with structured tool access, Claude Sonnet 4.6, without any fine-tuning, reached a similar error rate of 25.1 ± 1.1%, while Opus 4.8 reached 6.4%. This shows that introducing the tool can significantly reduce ($p < 0.0001$) the error rate across all LLMs. However, scientific instrument control requires a higher level of reliability. We therefore analyzed the failure profiles of the different models to identify the remaining sources of error.

Figure 2.c shows that although the overall error rates were similar for some configurations, the models failed for different reasons (Figure 2b, Supplementary Data - Table S2 and Table S3). For FT-GPT and FT-GPT-FC, 79.3% and 78.2% of errors, respectively, were related to incorrect units or values. In contrast, unit and value errors represented only 1.8% of the remaining errors for Sonnet 4.6 with tool access, where ambiguity or missing information accounted for 64.9%. For Opus 4.8, the remaining errors were mainly related to tool or module selection (64.3%), followed by ambiguity (21.4%).

Based on these failure modes, we introduced an ambiguity check layer before execution. The ambiguity check layer evaluates whether the user prompt contains sufficient information for a safe action and detects missing values or units, unclear axis directions, ambiguous tool targets, and unclear user intent (Supplementary Data – Table S4). Adding the ambiguity check layer reduced the error rate from 23.3 ± 0.7% to 8.9% for FT-GPT-FC ($P < 0.0001$), from 25.1 ± 1.1% to 4.8% for Sonnet 4.6 ($P < 0.0001$), and from 6.4% to 2.3% for Opus 4.8 ($P < 0.01$, Figure 2b). The reduction was larger than the fraction of errors initially classified as ambiguity. This indicates that under-specified requests can also lead to other error types, particularly incorrect units or values. Detecting missing information before execution therefore prevents errors.

We later implemented this tool interface through MCP (Supplementary Data - Information S1), which provides a modular connection between the language model and the instrument and supports an interactive workflow in which the human remains in the loop to clarify under-specified prompts before execution (Figure 2a). We show representative examples of both an interactive clarification case and a non-interactive case in which an ambiguous request is handled. In our benchmark, we evaluated FT-GPT-FC with the ambiguity check layer in the non-interactive form, whereas the Claude-MCP implementation allowed missing information to be provided within the same workflow before execution continued. The same type of interactive clarification could also be implemented with GPT-based systems.

We later implemented the tool interface through MCP, which provides a modular framework for connecting the language model to the instrument (Figure 2a). Within this framework, we added an interactive ambiguity-resolution step that keeps the human in the loop when a request is under-specified. We show representative examples of both interactive and non-interactive ambiguity handling. In our benchmark, FT-GPT-FC with the ambiguity check layer detected and withheld ambiguous requests, whereas the Claude-MCP implementation additionally allowed the operator to provide the missing information and continue execution within the same workflow. This interactive clarification is not specific to MCP and could also be implemented with GPT-based systems. The command profile of Claude MCP in Figure 2.d shows that of the 150 requests, 73.3% were executed directly, 22.7% required clarification, and 4.0% requested operations for which no tool was available.

Together, these results show that structured tool access reduced command errors, while ambiguity checking and interactive clarification further reduced the remaining errors before execution. The final interactive MCP configuration achieved the highest reliability among the evaluated systems, with no incorrect command execution observed in the benchmark.

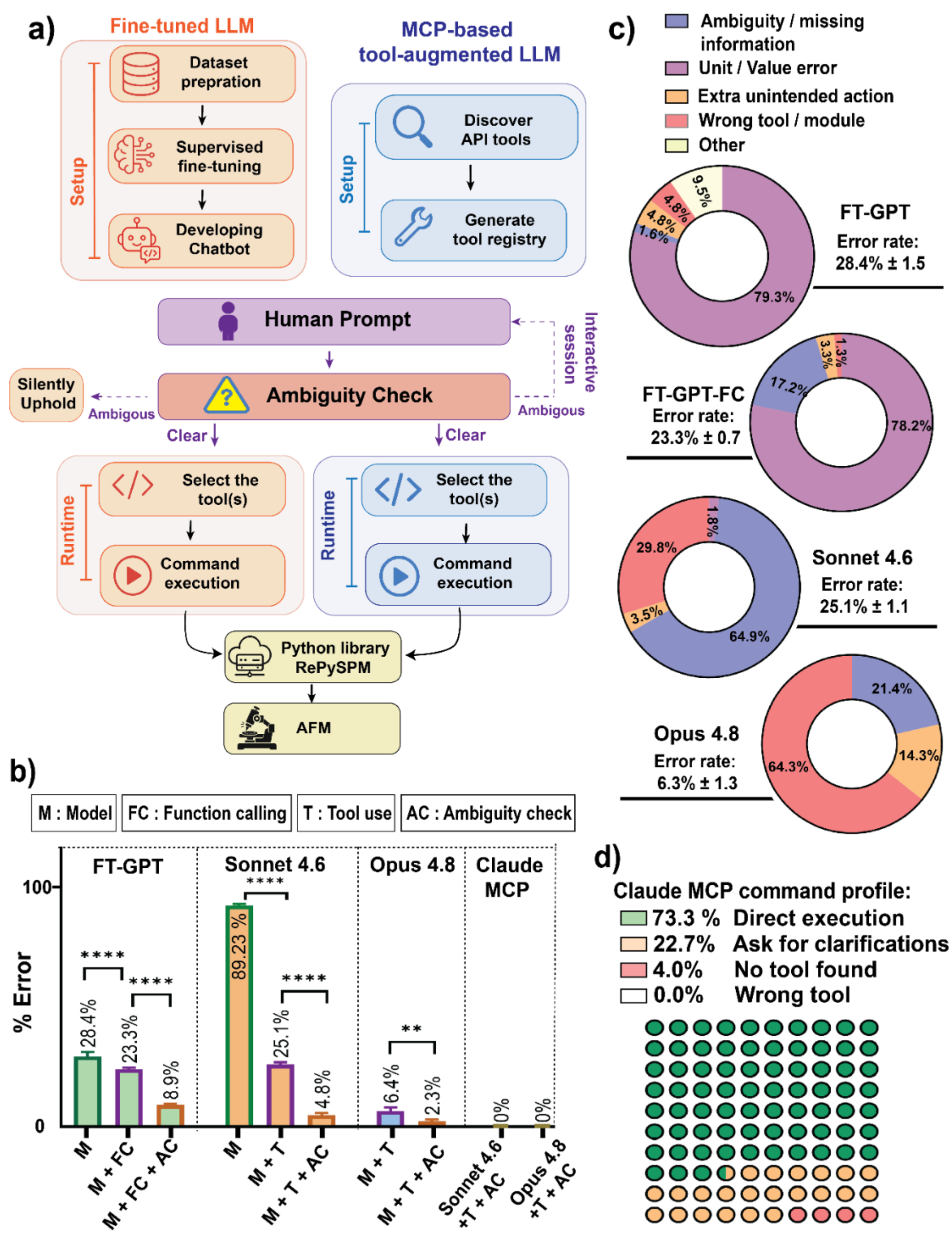

**Figure 2. Interactive ambiguity checking improves reliable natural-language control of an AFM. (a) Overview of the evaluated control workflows. In the fine-tuned workflow, GPT-4.1 was trained on instrument-specific command examples and evaluated with either direct command**

**generation (FT-GPT) or function calling (FT-GPT-FC). In the tool-based workflow, Claude Sonnet 4.6 and Claude Opus 4.8 were given structured access to 129 tools. The tool interface was subsequently implemented through MCP for the deployed interactive agent. Ambiguity checking (AC) identifies under-specified requests before execution. In the non-interactive configuration, flagged commands are withheld, whereas the interactive MCP workflow allows the operator to provide the missing information before execution continues. All workflows ultimately execute commands through RePySPM. b) Total command error rates for the evaluated configurations, with and without the ambiguity check layer. Structured tool access reduced command errors compared with the unaided model ($P < 0.0001$), while the ambiguity check layer further reduced the remaining errors for FT-GPT-FC and Sonnet 4.6 ($P < 0.0001$) as well as Opus 4.8 ($P < 0.01$). In the deployed interactive Claude-MCP configuration, we observed no incorrect command execution in the evaluated benchmark. Bars show mean ± SD across three independent runs. c) Error composition of the evaluated configurations without ambiguity check. Incorrect units or values dominated FT-GPT and FT-GPT-FC errors, whereas the remaining errors for tool-using Claude models were mainly associated with ambiguity, missing information, or incorrect tool or module selection. The total error rate for each configuration is shown alongside. d) Command profile of the deployed interactive Claude-MCP agent. Commands were either executed directly, returned for clarification, reported as having no matching RePySPM tool, or routed to an incorrect tool. Each circle represents 1% of evaluated commands.**

## 2.3 *AFM Pilot* takes control after safety validation

After validating the execution layer, we tested whether *AFM Pilot* could use the image and traces to tune the parameters from deliberately detuned starting conditions. *AFM Pilot* diagnoses imaging artifacts such as parachuting and feedback ringing and uses these observations to guide bounded parameter updates that improve surface-tracking quality while maintaining feedback stability. The tuning problem involves competing and coupled parameters. For example, increasing the integral gain can improve surface tracking but may introduce feedback ringing, while adjustment of the setpoint or scan rate may also be required depending on the current imaging state. The setpoint cannot be changed without limits, as maintaining a safe operating range is necessary to safeguard the tip from damage, while reducing the scan rate can improve tracking at the cost of acquisition speed. The optimal parameter update therefore depends on both the observed imaging problem and the current values of the other parameters. In our framework, we provide these relationships and safety constraints to the LLM as operating instructions, allowing it to select among bounded parameter changes based on the current imaging state.

Figure 3a shows closed-loop parameter tuning while imaging a standard calibration grating. The initial scan starts from a non-tracking condition with strong structure in the error channel (Iter. 0). Across successive iterations, *AFM Pilot* adjusts the imaging parameters and progressively improves surface tracking and topographic fidelity. As tracking improves, parachuting becomes less pronounced, and the error signal decreases, while feedback stability is maintained without excessive ringing. This result shows that the agent can recover a stable, well-tracking image from a severely non-optimal starting point. The tuning is not restricted to PID adjustment; when necessary, *AFM Pilot* can jointly adjust imaging parameters such as integral gain, setpoint, and scan rate within predefined bounds, similar to the iterative tuning performed by a human AFM operator (Supplementary Video S1).

We next tested whether the same image-driven tuning strategy works on biological samples with more complex structure (Figure 3b and Supplementary Data - Figure S1). Figure 3b.I shows the sequence of a butterfly wing tuned and imaged by *AFM Pilot*. Next to it is an image of the same area with imaging parameters pre-tuned by a human expert on the same feature. *AFM Pilot* starts from detuned parameters, where the surface is not properly tracked, and iteratively assesses forward and backward height and error data through *AI vision*. In the next panel (Figure 3b.II), we show the data sent to the LLM, including the forward and backward height and error channels acquired since the most recent parameter update, along with the exact traces of the last ten forward and backward scan lines used for assessment. It then adjusts parameters such as the setpoint, feedback gain, and scan rate within predefined bounds to improve surface tracking and maintain feedback stability. The presence and severity of parachuting and feedback ringing are used to assess whether further tuning is required. The tuning continues until the agent judges that the image is "as good as it gets" within the allowed parameter space. For artifacts outside this control space, such as hysteresis, creep, or double-tip effects, it reports the likely cause and suggests the appropriate operator-level intervention.

We also applied the same tuning strategy to a more advanced AFM imaging mode, Kelvin-probe force microscopy (KPFM), using a gold-aluminum grating sample (Figure 3c). *AFM Pilot* recovered the topography with feature quality comparable to human operation. During this experiment, the KPFM-specific PID gain was kept fixed, while the acquired KPFM channel was provided to *AFM Pilot* together with the topography and error information so that it could assess whether image quality remained acceptable across all channels. Together, these results show that *AFM Pilot* can be applied with the same AI-based tuning strategy across sample types and imaging modes. A representative session report is provided in Supplementary Data - Information S3.

Finally, prompt caching substantially reduced repeated input processing during AFM tuning. Over a ten-iteration session, approximately 95% of processed input tokens were served from the cache. As the number of iterations increases, the number of cached tokens increases; however, the cost of using cached data is 0.1 of the basic prices.

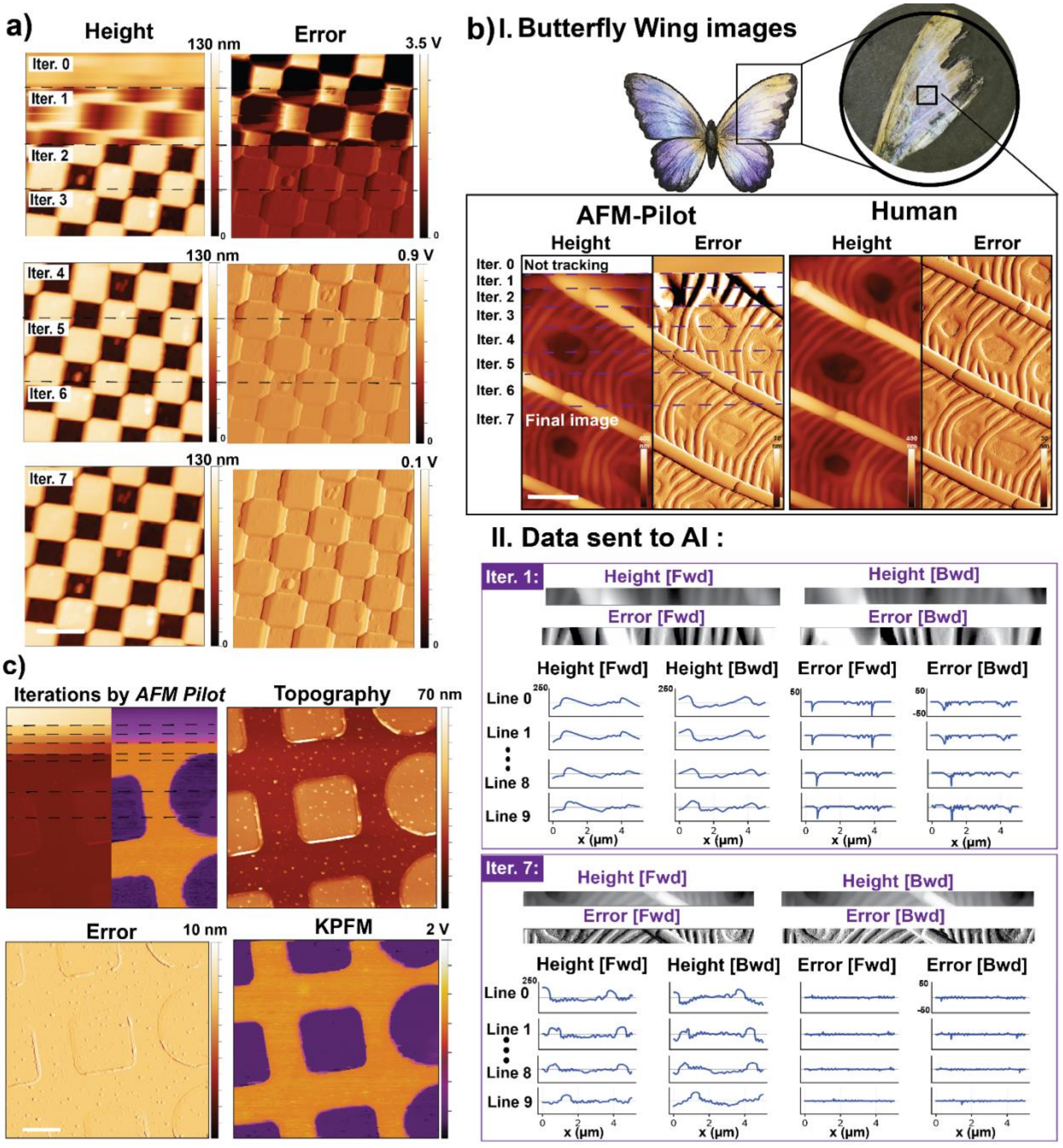

**Figure 3.** ***AFM Pilot*** **uses AI vision to tune AFM images across different samples and imaging modes. a) Calibration grating tuned over eight iterations from a non-tracking initial condition to a stable final image. The dashed lines indicate the parameter-update points during tuning.**

**As tracking improves, the checkerboard topography becomes sharper, and the error signal decreases. Scale bar = 200 nm. b) I. Butterfly wing imaged by *AFM Pilot* and by a human expert on the same feature. *AFM Pilot* starts with detuned parameters and iteratively adjusts the imaging parameters until it achieves stable tracking. The final height and error images are shown next to those acquired by the human operator. Scale bar= 1 µm. II. The panel illustrates the information provided to the agent during tuning: forward and backward height and error data acquired after the latest parameter update, together with the traces of the last ten lines forward and backward. These traces allow the agent to assess line-to-line tracking and evaluate artifacts such as parachuting or feedback ringing before selecting the next parameter update. The improvement in height and reduction of the errors are comparable in iteration 1 to iteration 7. c) Extension of the same tuning strategy to KPFM on a gold–aluminum grating. The left-top image shows the successive tuning iterations, while the final topography, error, and KPFM channels are also shown. We kept the KPFM-specific PID gain fixed and provided the KPFM image to AFM Pilot along with the topography and error channels to assess image quality across all acquired signals. Scale bar = 10 µm.**

## 2.4 *AFM Pilot* matches human operators in live tuning

To confirm the performance of the agent, we performed a comparative test between five human operators and *AFM Pilot* under matched conditions, starting from the same detuned parameters and using the same image-quality criteria. For this comparison, the adjustable parameter was intentionally limited to setpoint and integral gain to enable a controlled and direct comparison between human and AI tuning strategies. Figure 4a shows the trajectories of these parameters as the human operators and *AFM Pilot* tuned the feedback to improve surface tracking while maintaining feedback stability.

The trajectories show that the agent made more conservative setpoint adjustments than the human operators, keeping the setpoint closer to its initial value, while the integral-gain changes were comparable. Both the human operators and *AFM Pilot* converged toward improved surface tracking while maintaining feedback stability. The final images also showed parachuting severity below the predefined acceptable threshold. This threshold was established through agreement between expert visual assessment and a parallel Python-based numerical detector used to quantify artifact severity, as described in Methods Section 5.5. Both also showed a transient increase in feedback ringing during tuning, particularly as integral gain increased, before reaching a stable final state, reflecting the expected trade-off between surface tracking and feedback stability.

Differences between human operators and *AFM Pilot* were evaluated using the Wilcoxon signed-rank test for paired observations ($n = 5$ matched pairs). Across the evaluated endpoints, no statistically significant paired differences were detected (Figure 4b). The number of iterations ($p = 0.063$), tuning time ($p = 0.625$), final parachuting severity ($p = 1.00$), and final ringing severity ($p = 0.188$) did not differ significantly between human operators and the *AFM Pilot*. So, the *AFM Pilot* required fewer iterations and reached a lower final ringing severity, whereas the human operators completed tuning slightly faster and reached a slightly lower final parachuting severity. Despite differences in the individual tuning trajectories, both reached acceptable final imaging states under identical starting conditions, adjustable parameters, and safety bounds.

### 2.5 *AFM Doctor* diagnoses and treats image artifacts

Finally, *AFM Doctor* handles the post-processing step of the pipeline (Figure 4c). It inspects the acquired image through *AI vision*, identifies common artifacts, and explains their likely physical cause to the user. The agent then selects a correction from a pre-approved set of transparent processing tools, which are listed under the method section. Figure 4c shows a representative raw image with scars, height offsets, and background tilt. *AFM Doctor* diagnosed the problems and applied scar removal, background leveling, and line flattening, producing a result visually comparable to the image processed by a human expert. After each correction, the agent can re-inspect the processed image, allowing additional correction steps when necessary. A representative of a session report is provided in Supplementary Data - Information S5.

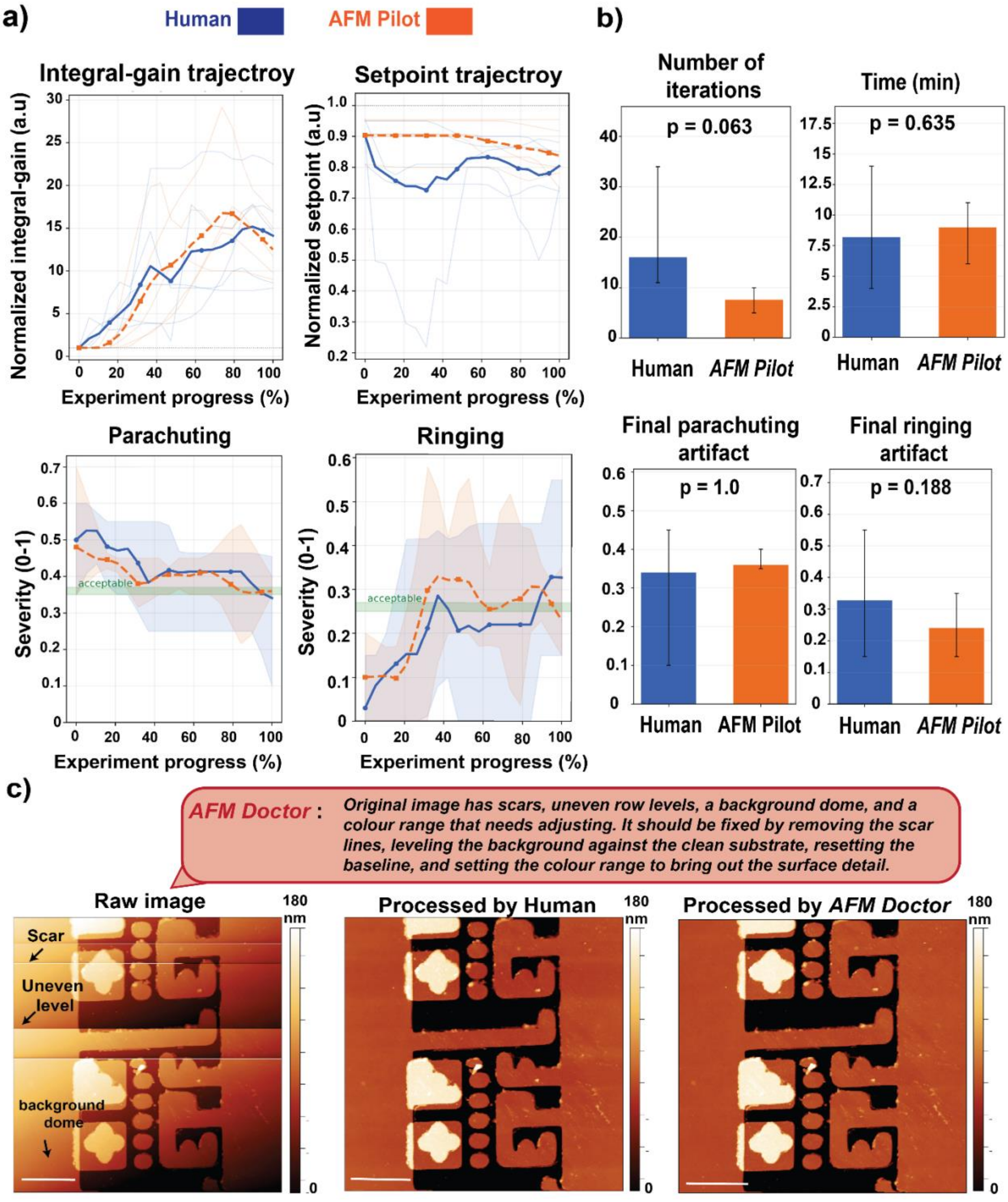

**Figure 4.** ***AFM Pilot*** **performs live parameter tuning relative to human experts, and** ***AFM Doctor*** **corrects image artifacts using pre-approved processing steps. a) Setpoint and integral-gain trajectories, together with parachuting and feedback-ringing severity, during tuning by** ***AFM Pilot*** **and human experts. Tuning progress is expressed as the percentage of total tuning progress for each session, with 0% corresponding to the initial detuned condition and 100% to**

**the final accepted image, allowing sessions of different lengths to be compared on the same relative axis. Integral gain is normalized to its initial value for each run, whereas setpoint is normalized to the free amplitude. Artifact severity is scored from 0 to 1 based on the *AI vision* assessment. For the parameter trajectories, faint lines represent individual sessions and bold lines represent the mean. For the artifact trajectories, bold lines represent the mean and shaded regions indicate the range across sessions. *AFM Pilot* follows a similar integral-gain trajectory to the human experts while making more conservative setpoint adjustments. Both *AFM Pilot* and the human experts converge toward improved surface tracking and stable feedback, with final parachuting and ringing severities within the predefined acceptable ranges. b) Summary endpoints for the number of iterations, tuning time, final parachuting severity, and final ringing severity. Differences between human operators and *AFM Pilot* were evaluated using a two-sided Wilcoxon signed-rank test for paired observations ($n = 5$ matched pairs). No statistically significant paired differences were detected for the number of iterations ($p = 0.063$), tuning time ($p = 0.625$), final parachuting severity ($p = 1.000$), or final ringing severity ($p = 0.188$). Bars show the mean, with error bars indicating the range across the five sessions. c) *AFM Doctor* post-processing. A raw image containing scars, background tilt, and regions requiring cropping is compared with the corresponding images processed by a human expert and by *AFM Doctor* using pre-approved operations, including scar removal and background leveling.**

# 3. Discussion

This work brings together three parts of AFM operation that are usually treated separately; A safe command execution, image-based parameter tuning, and post-processing. We show that a general-purpose, tool-augmented LLM can operate the executable part of an AFM workflow without task-specific training. The framework combines three MCP-based agents. *AFM Messenger* translates natural-language instructions into safe instrument commands, *AFM Pilot* assesses acquired AFM images and selects bounded parameter updates, and *AFM Doctor* applies transparent post-processing steps from a pre-approved tool set. This separation keeps image-based reasoning and hardware execution connected within the same workflow while preventing uncertain or under-specified commands from reaching the microscope.

The main contribution is not simply connecting an LLM to an AFM, but enabling the agent to assess the measurement itself and decide which operational action is needed next. Conventional autonomous methods usually depend on a predefined objective, such as a reward function[20], tracking metric[42], or trace-retrace similarity score[39]. Although such objectives work effectively, they only capture the features they were designed to measure. This is also the key difference from prior LLM-operated AFM, where the LLM coordinated the workflow while imaging parameter optimization was performed externally by an external operator such as a genetic algorithm[32,39]. Here, *AFM Pilot* uses *AI vision* to assess the height image, error image, scan-line information, and artifact severity together. Using the general scientific and visual knowledge of the LLM, without training, it identifies the likely imaging problem, scores its severity, and selects the parameter that should be changed. In this way, the LLM moves from workflow coordination toward direct instrument operation, closer to how an experienced AFM operator analyzes a scan and adjusts parameters during acquisition.

This approach complements ongoing work on autonomous SPM. Active-learning approaches have enabled microscopes to autonomously select informative measurement locations or explore structure-property relationships according to a defined acquisition function or reward[18,19,43]. Hypothesis-learning approaches extend this concept by allowing autonomous SPM experiments to select measurements that distinguish between competing physical models[44]. More recently, reward-driven tuning has been used to optimize SPM imaging parameters from an explicitly defined image-quality reward[20]. Other autonomous SPM approaches have focused on automated image-quality assessment, probe conditioning, and region selection[45], and AFM operation using task-specific machine-learning models[21,46]. These approaches are powerful when the experimental objective can be expressed quantitatively. In contrast, *AFM Pilot* uses the LLM to interpret several aspects of image quality together before selecting an operational action. The approaches are therefore complementary; the numerical optimization and active learning can efficiently search a defined parameter or measurement space, while LLM-based reasoning may be useful when the decision depends on several image features that are difficult to combine into a single objective.

Our experiments further showed why the architecture around the model matters for accurate command execution during controlling physical instruments. Different LLMs showed different types of errors. Structured tool access reduced the overall error rate, while the ambiguity check layer safeguarded against executing under-specified or unclear prompts. This is important because an instrument command can be syntactically valid while still containing a physically inappropriate value. The role of the ambiguity check layer is therefore not to make the LLM itself error-free, but to prevent uncertainty or misinterpretation from becoming a hardware action.

We also showed that an interactive workflow is important for keeping the human in the loop when clarification is required. The role of human intervention in handling uncertainty has also been discussed before in autonomous SPM[47]. In the Claude-MCP tests, allowing the operator to provide missing information enabled ambiguous requests to be resolved before execution. Interactive clarification is not specific to MCP and could also be implemented with other LLM architectures. MCP was useful in our implementation because the tool registries can be modified independently of the language model. It also allows the three agents to operate with specialized tool sets within the same workflow, while the LLM orchestrates their interactions and selects the appropriate agent for each task. In addition, the Claude-MCP client maintains the interaction history within a session (i.e., memory handling), which is useful for longer experiments because previous commands, clarifications, and intermediate decisions remain available as context.

The comparison with human experts suggests a practical role for *AFM Pilot* as an expert assistant rather than a replacement for experienced operators. It could be particularly useful in research groups where AFM is used as a supporting technique by biologists, materials scientists, or other researchers who need reliable imaging without developing deep expertise in instrument-specific parameter tuning. In addition, we noticed that human users combine the acquired image with prior knowledge of the sample, probe, imaging mode, expected morphology, and previous measurements. Previous autonomous SPM studies have also shown that prior knowledge about the sample or underlying physics can be used to guide automated measurement decisions [43,44]. We observed similar behavior in our *AFM-Pilot* agent when it was informed about a sample (e.g., collagen fibers) and considered the possibility that its characteristic periodic banding could interfere with the detection of feedback ringing (Supplementary Data – Information S4).

Here, we intentionally limited the adjustable parameters to setpoint and integral gain to enable controlled tuning and direct comparison with human experts. The framework could be extended to a broader parameter set, provided that parameter interactions and safe operating ranges are carefully defined. Expanding the parameter scope also raises the question of which decisions an agent should handle and which should remain under conventional control. Agent-based control is particularly useful when image quality depends on competing parameters and cannot be represented by a single analytical objective. However, classical or optimal control remains more suitable when system dynamics are well characterized and rapid deterministic responses are required[48–51]. Recent reward-driven SPM tuning demonstrates that numerical optimization can already automate parameter adjustment when an appropriate reward can be defined[20]. A hybrid architecture could combine conventional low-level control with agent-based image interpretation and higher-level tuning decisions similar to what Cissé et al. 52 proposed, but with an extended range of parameters.

The current implementation also introduces timing and resource constraints. Each iteration requires data acquisition after a parameter change, followed by transfer to the LLM and model inference. Response time varies with internet connection and external model availability, making the tuning duration less predictable. Repeated transmission of image data also increases token consumption and operating costs. A locally deployed model in future work could reduce these constraints, although its image assessment, command reliability, and ambiguity handling would require validation.

Another opportunity for further study is to provide *AFM Pilot* with information from more imaging channels. The present framework uses height and error, which are broadly available across AFM modes. However, tapping-mode AFM can provide additional signals such as phase, and advanced electrical or mechanical modes can produce several simultaneous channels. Kalinin et al. have already demonstrated automated PFM experiments in which the system learns which of several imaging channels is most predictive of the property of interest[52]. Such multiparametric data may be particularly interesting for an LLM-based agent because the model could consider relationships between several channels when assessing image quality. This could become particularly useful when the number of coupled signals becomes difficult for a human operator to evaluate simultaneously. Whether an LLM can actually use this additional information better than an experienced operator, however, remains an open question and should be tested directly. Additional channels do not necessarily improve performance and may provide redundant or conflicting information. Future work should therefore test whether selected channel combinations improve tuning compared with height and error alone.

Finally, the experimental intent remains human-defined. The system can translate prompts, tune parameters, and post-process images, but it does not yet decide what experiment should be performed. Other autonomous SPM approaches are already addressing parts of this higher-level decision process. A future system could combine these capabilities, with active-learning[18,19] or hypothesis-driven methods[44] deciding what or where to measure, and an agentic control layer deciding how to execute and adapt the measurement safely. We therefore frame this work as agentic operation of a scientific instrument rather than as a fully self-driving laboratory.

## 4. Conclusion

We show that a general-purpose, tool-augmented LLM can operate the executable part of an AFM workflow without AFM-specific retraining. Through three MCP-based agents, *AFM Messenger* translates natural language instructions into validated instrument commands, *AFM Pilot* evaluates acquired images and selects bounded parameter adjustments, and *AFM Doctor* applies transparent post-processing from a pre-approved tool set. Safe operation depends not only on model capability, but also on structured tools, unit validation, ambiguity handling, and clarification before hardware execution. In live experiments, *AFM Pilot* achieved performance comparable to AFM operators under matched detuned conditions. More broadly, this work presents an approach to scientific instrument automation in which AI agents operate through established procedures, tuning strategies, and safeguards developed through decades of human expertise, rather than requiring the instrument workflow to be redesigned around machine-specific objectives and assessment metrics.

## Acknowledgements

This project received funding from the European Union's Horizon 2020 research and innovation program under the Marie Skłodowska-Curie grant agreement No 945363 (**EPFLGlobaLeaders**,

G.E.F., Z.A). This work was also supported by **EPFL Center for Imaging** under grant No. 563292 (G.E.F. and Z.A), the Swiss National Science Foundation (**SNSF**, Video-rate nanomechanical properties mapping using atomic force microscopy, grant **No**. 200021_175675; G.E.F. and M.P.; **MEFS**, grant **No**.200021_182562, **No**.200020_213072, and Our research on artificial Ni81Fe19 quasicrystals on **No**.163016; G.E.F., M.P., and N.H.), the European Research Council (ERC-2017-CoG, **InCell**, No.773091; G.E.F. and M.P.), the European Union funding (Framework Programme FP7/2007-2013/ERC under Grant Agreement No. 307338-**Eurostars E**! 8213-Triple-S, G.E.F., M.P and N.H), the **Innosuisse** - Swiss Innovation Agency (**AFM with PORT**: Atomic force microscope with photothermal off-resonance tapping, No.36938.1 IP-EE; G.E.F. and M.P.; **NAFTAQ**: Novel AFM Techniques for Autonomous Quality Control in Industrial Manufacturing, 102.271 IP-EE, No.10632; G.E.F., P.S., M.P., and M.M.; E!9399 **TOPFA**: Top-Side Electrical Failure Analysis for Advanced Semiconductor Nodes, No.11931; G.E.F., P.S. and M.M.), the **ETH Domain ORD** Program (Open SPM, No.10126, and Open SPM+, No.ORD200165; G.E.F. and P.P.S., and M.P.), and the Technology Agency of the Czech Republic (**IM Beast**, No.TACR FW10010168; G.E.F., M.P. and P.P.S. The authors thank the contributors to the RePySPM open-source project.

## Data and Code Availability

The RePySPM instrument controller is available in GitHub repository [40]. The MCP server implementations for *AFM Messenger*, *AFM Pilot*, and AFM Doctor, the benchmark datasets, fine-tuning scripts, and evaluation code will be made available as an open-source codebase upon publication[41].

# 5. Methods

### 5.1 Instrument and software interface

Experiments were performed on a custom open-hardware AFM[53–56] platform controlled through RePySPM[40], an open-source Python API that enables communication with the custom microscope control software written in Python. RePySPM provides high-level access to instrument functions organized into modules for z-control, scan-parameter configuration, scan control, motors, signals, lasers, image acquisition, and utility operations.

### 5.2 Data preparation

An experienced AFM user wrote a command list of 234 prompts/commands, then augmented it by pairing natural-language operator instructions with the corresponding RePySPM Python function calls. An experienced AFM user screened all scenarios to create a dataset of 2747 scenarios (prompt + expected answer) (datasets are provided in GitHub repository[41]). The dataset combined real operator sessions with synthetic examples covering rarely used instrument functions and was randomly divided into training (76%), validation (8%), and test sets (two sets of 8%). We used three fixed datasets for all tests in all models, including: 1) dataset 1 (containing 225 test cases), dataset 2 (containing 200 randomly shuffled scenarios sampled from dataset 1 using a fixed random seed (seed = 42)), and dataset 3 (224 test cases). All data are reported as mean ± standard deviation across these three datasets.

### 5.3 Fine-tuned text-generation system

The model was fine-tuned based on the base model gpt-4.1-2025-04-14 (batch size =3, epochs=3, LR multiplier= 2) using the mentioned training and validation datasets. Each example was formatted as a short system, user, and assistant exchange. During inference, the fine-tuned model received the same system prompt and generated Python code as free text at temperature 0, without schema constraints or output validation.

As an additional baseline (FT-GPT-FC), we evaluated the same fine-tuned model with function calling on the tool schema, without further training and without the interactive ambiguity check. With the model held constant, this comparison estimates the benefit of structured tool selection alone; the remaining errors define the gap the ambiguity check layer is designed to close.

A lightweight chatbot, implemented with Flask and an HTML front end, was developed to connect user commands to the GPT model and execute the generated code.

### 5.4 *AFM Messenger* for safe command execution and ambiguity checking

*AFM Messenger* was implemented as the MCP-based agent responsible for translating natural-language instructions into safe instrument commands. A discovery script scanned the RePySPM API library and converted each public method into an MCP tool containing the module name, function name, parameters, and unit information. The resulting tools were stored in a JSON registry and loaded through FastMCP[57]. When RePySPM changes, the registry can be regenerated by re-running the discovery script without retraining the model. In total, 128 tools were exposed and individually validated with scripted test cases before autonomous operation.

*AFM Messenger* included an ambiguity check layer designed from the observed error taxonomy of the fine-tuned and Claude models. The rules distinguish SET and GET operations, route shared parameters such as setpoint to the correct module, enforce unit conversions, and separate parameter configuration from scan execution. Prompts with missing values, uncertain units, unclear axes, or ambiguous targets were returned to the operator for clarification before any tool was invoked. Safety-critical actions, including approach and scan initiation, required an additional confirmation (Supplementary Material S1).

For benchmarking, each system was evaluated on the three mentioned datasets in Method 5.2. The datasets covered the tools in *AFM Messenger* and included both complete and underspecified commands. We assessed correctness by exact match between the expected function call or tool selection and the model output at temperature 0. We manually classified failures as unit or value errors, ambiguity or missing information, extra unintended action, wrong tool or module, or other errors. We computed means and standard deviations across the three datasets, and report full results in Supplementary Tables S1–S3. We evaluated Claude-MCP with Claude Sonnet 4.6 and Claude Opus 4.8 and compared it with the same Claude models without the ambiguity check layer, as well as fine-tuned GPT with and without function calling.

### 5.5 *AFM Pilot* for closed-loop imaging

*AFM Pilot* was implemented as the MCP-based agent responsible for image-based parameter tuning. At each iteration, *AFM Pilot* retrieved the most recent forward and backward height and error data through *AFM Messenger*. The agent rendered these data as diagnostic panels together with the current setpoint, integral gain, and scan rate. Using *AI vision*, the model assessed line-to-line surface tracking and feedback stability and scored six imaging problems, including parachuting, feedback ringing, hysteresis, creep, double-tip effects, and loss of surface tracking. We used these assessments to guide bounded parameter updates aimed at improving surface tracking while maintaining feedback stability. For the tuning experiments reported here, the model returned a severity score, a short visual rationale, and bounded updates to the permitted imaging parameters. The severity score is a judgment made by the model rather than a computed quantity. Its meaning is fixed by verbal anchors defined in the agent prompt, where 0.0 indicates that the artifact is absent, 0.1–0.3 a subtle presence, 0.4–0.6 a clearly visible artifact, and 0.7–1.0 a severe artifact. Expert visual assessment of image quality, rather than a numerical optimization criterion, established the acceptable severity ranges. Parachuting severity $\leq 0.35$ and ringing severity $\leq 0.25$ were considered acceptable, corresponding to residual artifact levels that an experienced operator judged would not compromise the quality of the acquired image.

Operator-defined safety bounds, including scan-rate limits, a minimum setpoint, and a parameter-priority order, constrained the closed-loop tuning process. At each iteration, observed parachuting and ringing severity, along with height and error data, indicated whether further tuning was required. The loop continued until surface tracking and feedback stability reached the operator-defined acceptable state, or until the model determined that no further beneficial parameter adjustment could be made within the allowed bounds. The final accepted image was selected from the tuning sequence based on the artifact-severity assessment and overall visual image quality. Supplementary Data – Information S3 shows a conversation from a representative *AFM Pilot* session. Prompt caching was applied to the fixed *AFM Pilot* vision instructions. The system prompt containing artifact-scoring criteria, tracking rules, and the required response format was cached, whereas iteration-specific inputs, including the composite image, line profile, and current instrument state, were provided at each iteration. Because iterations were separated by approximately ten seconds, the cached prompt remained available throughout a typical tuning session.

In parallel with the *AI-vision* assessment, a deterministic detector implemented in Python computes an independent severity score between 0 and 1 for three of the six artifact classes, directly from the forward and backward height and error line arrays. The deterministic detector does not participate in the tuning loop or determine parameter updates; instead, it provides an independent quantitative cross-check of the *AI-vision* assessment. The measures introduced here follow established principles, but their numerical form is specific to this work. Parachuting is identified from a threshold on the local gradient at feature edges, combined with the asymmetry between the forward and backward traces, a principle introduced by Kubo et al. for two-way scanning data[58]. In our implementation, the error signal of each line is compared with a robust baseline formed from the median and the median absolute deviation; sustained deviations entered through a sharp edge are marked, and the severity is taken as the fraction of the line they occupy, combined with the residual between the forward and backward height traces near feature edges. Feedback ringing is identified from the spectral content of the tracking signal, following the principle used by Kohl et al. for automatic gain tuning, where ringing appears once the feedback gains exceed the stability margin[59]. We take the fraction of power above one-fifth of the Nyquist frequency in the line-averaged power spectrum and normalize it between anchor values measured on the same scanner from a clean frame and from a frame recorded with deliberately excessive gain. Hysteresis is measured by cross-correlating each forward line with its backward counterpart and taking the lag of the correlation maximum as the spatial offset between the two scan directions, following the correlation method used for scanner distortion by Chu et al.[60], applied here to whole lines to obtain a single severity value rather than in moving windows to build a correction map. The offset is normalized to the line width and weighted by its consistency across lines**.** These deterministic scores run alongside, but independently of, the AI-vision assessment. The AI-vision assessment guides parameter updates, whereas the deterministic detector provides a transparent, reproducible cross-check (Supplementary Data – Information S3). Once the difference between the AI- and Python-based assessments exceeds 50%, it triggers the LLM to recheck the data.

### 5.6 *AFM Doctor* for artifact diagnosis and post-processing

*AFM Doctor* was implemented as the MCP-based agent responsible for artifact diagnosis and post-processing. The agent loaded acquired images and accessed a fixed, pre-approved set of processing tools, including plane leveling, row alignment, scar removal, cropping, background subtraction, and file export. All code is provided in GitHub repository[41]. At each step, the image was rendered for *AI-vision* assessment. The model identified remaining artifacts, explained the likely physical cause, and selected an appropriate correction from the approved tool set.

Restricting *AFM Doctor* to transparent processing tools prevented unrestricted image manipulation and reduced the risk of introducing unsupported artifacts. After processing, the agent could re-inspect the image to determine whether further approved corrections were needed. Human reference processing was performed in Gwyddion[61] open-source software.

AFM Doctor provides a set of established SPM correction operations, including line flattening63, background leveling63, scar removal63, feature-masked substrate-only leveling63, substrate-anchored multi-point leveling63, directional FFT destriping64, baseline zeroing64, and color-range setting[62]. Following established practice, the line offsets and polynomial backgrounds are derived only from the masked flat-background region rather than from the sample topography, so corrections are not biased by surface features. Each operation is paired with AI-vision artifact diagnosis and an objective background-flatness metric, and results are exported as annotated, figure-ready images and editable Gwyddion files.

### 5.6 Human-operator comparison

For the expert comparison, five human operators and *AFM Pilot* began from matched detuned conditions, used the same adjustable parameters and safety bounds, and were evaluated using the same image-quality measures. Based on visual assessment by experienced AFM users, parachuting severity scores of approximately 0.30–0.35 and ringing severity scores of approximately 0.20–0.25 were considered acceptable. For *AFM Pilot*, reaching these ranges indicated that further tuning was not required unless another aspect of image quality remained suboptimal. Human operators stopped tuning when they judged the image acceptable based on their own visual assessment; parachuting and ringing severity were then tracked throughout each session using the same scoring framework. Control-parameter and artifact trajectories were normalized to tuning progress, defined as the percentage of total tuning time for each session, allowing sessions of different durations to be compared on a common relative axis. The comparison endpoints were the number of iterations, tuning time, final parachuting severity, and final ringing severity.

### 5.7 Statistical Analysis

Data analysis was performed using GraphPad Prism and the accompanying Python analysis layer. Results are reported as mean ± standard deviation unless otherwise stated. Group differences were assessed using one-way ANOVA followed by Tukey post hoc multiple-comparison tests. SPM visualization and processing with human were performed using Gwyddion.

## References

1. Wang, H. *et al.* Scientific discovery in the age of artificial intelligence. *Nature* **620**, 47–60 (2023).
2. Abou Ali, M., Dornaika, F. & Charafeddine, J. Agentic AI: a comprehensive survey of architectures, applications, and future directions. *Artif. Intell. Rev.* **59**, 11 (2025).
3. Acharya, D. B., Kuppan, K. & Divya, B. Agentic AI: Autonomous Intelligence for Complex Goals—A Comprehensive Survey. *IEEE Access* **13**, 18912–18936 (2025).
4. Hartung, T. AI, agentic models and lab automation for scientific discovery — the beginning of scAInce. *Front. Artif. Intell.* **8**, (2025).
5. Tom, G. *et al.* Self-Driving Laboratories for Chemistry and Materials Science. *Chem. Rev.* **124**, 9633–9732 (2024).
6. M. Bran, A. *et al.* Augmenting large language models with chemistry tools. *Nat. Mach. Intell.* **6**, 525–535 (2024).
7. Szymanski, N. J. *et al.* An autonomous laboratory for the accelerated synthesis of inorganic materials. *Nature* **624**, 86–91 (2023).
8. Li, C., Ran, N. & Liu, J. Agentic material science. *J. Mater. Inform.* **6**, N/A-N/A (2026).
9. Abolhasani, M. & Kumacheva, E. The rise of self-driving labs in chemical and materials sciences. *Nat. Synth.* **2**, 483–492 (2023).
10. Delgado-Licona, F. & Abolhasani, M. Research Acceleration in Self-Driving Labs: Technological Roadmap toward Accelerated Materials and Molecular Discovery. *Adv. Intell. Syst.* **5**, 2200331 (2023).
11. Yoshikawa, N. *et al.* Large language models for chemistry robotics. *Auton. Robots* **47**, 1057–1086 (2023).

12. Darvish, K. *et al.* ORGANA: A robotic assistant for automated chemistry experimentation and characterization. *Matter* **8**, (2025).
13. Raptis, E. K., Kapoutsis, A. C. C. & Kosmatopoulos, E. B. Agentic LLM-based robotic systems for real-world applications: a review on their agenticness and ethics. *Front. Robot. AI* **12**, (2025).
14. Vriza, A., Prince, M. H., Zhou, T., Chan, H. & Cherukara, M. J. Operating advanced scientific instruments with AI agents that learn on the job. *Npj Comput. Mater.* **12**, 160 (2026).
15. Durand, A. *et al.* A machine learning approach for online automated optimization of super-resolution optical microscopy. *Nat. Commun.* **9**, 5247 (2018).
16. Morgado, L., Gómez-de-Mariscal, E., Heil, H. S. & Henriques, R. The rise of data-driven microscopy powered by machine learning. *J. Microsc.* **295**, 85–92 (2024).
17. Kousaka, J., Iwane, A. H. & Togashi, Y. Automated cell structure extraction for 3D electron microscopy by deep learning. *Sci. Rep.* **15**, 17481 (2025).
18. Kalinin, S. V. *et al.* Automated and Autonomous Experiments in Electron and Scanning Probe Microscopy. *ACS Nano* **15**, 12604–12627 (2021).
19. Vasudevan, R. K. *et al.* Autonomous Experiments in Scanning Probe Microscopy and Spectroscopy: Choosing Where to Explore Polarization Dynamics in Ferroelectrics. *ACS Nano* **15**, 11253–11262 (2021).
20. Liu, Y. *et al.* Machine Learning-Based Reward-Driven Tuning of Scanning Probe Microscopy: Toward Fully Automated Microscopy. *ACS Nano* **19**, 19659–19669 (2025).
21. Kang, S., Park, J. & Lee, M. Machine learning-enabled autonomous operation for atomic force microscopes. *Rev. Sci. Instrum.* **94**, 123704 (2023).
22. Ziatdinov, M., Ghosh, A., Wong, C. Y. (Tommy) & Kalinin, S. V. AtomAI framework for deep learning analysis of image and spectroscopy data in electron and scanning probe microscopy. *Nat. Mach. Intell.* **4**, 1101–1112 (2022).

23. Binnig, G., Quate, C. F. & Gerber, Ch. Atomic Force Microscope. *Phys. Rev. Lett.* **56**, 930–933 (1986).
24. Kodera, N., Sakashita, M. & Ando, T. Dynamic proportional-integral-differential controller for high-speed atomic force microscopy. *Rev. Sci. Instrum.* **77**, 083704 (2006).
25. Stirling, J. Control theory for scanning probe microscopy revisited. *Beilstein J. Nanotechnol.* **5**, 337–345 (2014).
26. Habibullah, Pota, H. R., Petersen, I. R. & Rana, M. S. Creep, Hysteresis, and Cross-Coupling Reduction in the High-Precision Positioning of the Piezoelectric Scanner Stage of an Atomic Force Microscope. *IEEE Trans. Nanotechnol.* **12**, 1125–1134 (2013).
27. Han, C. & Chung, C. C. Reconstruction of a scanned topographic image distorted by the creep effect of a *Z* scanner in atomic force microscopy. *Rev. Sci. Instrum.* **82**, 053709 (2011).
28. Matsunaga, Y., Fuchigami, S., Ogane, T. & Takada, S. End-to-end differentiable blind tip reconstruction for noisy atomic force microscopy images. *Sci. Rep.* **13**, 129 (2023).
29. Bian, K. *et al.* Scanning probe microscopy. *Nat. Rev. Methods Primer* **1**, 36 (2021).
30. Burnham, N. A., Lyu, L. & Poulikakos, L. Towards artefact-free AFM image presentation and interpretation. *J. Microsc.* **291**, 163–176 (2023).
31. Arias, S., Zhang, Y., Zahl, P. & Hollen, S. Autonomous Molecular Structure Imaging with High-Resolution Atomic Force Microscopy for Molecular Mixture Discovery. *J. Phys. Chem. A* **127**, 6116–6122 (2023).
32. Liu, Y., Checa, M. & Vasudevan, R. K. Synergizing human expertise and AI efficiency with language model for microscopy operation and automated experiment design. *Mach. Learn. Sci. Technol.* **5**, (2024).
33. Anisuzzaman, D. M., Malins, J. G., Friedman, P. A. & Attia, Z. I. Fine-Tuning Large Language Models for Specialized Use Cases. *Mayo Clin. Proc. Digit. Health* **3**, 100184 (2025).

34. Hu, Y., Kim, H., Ye, K. & Lu, N. Applying fine-tuned LLMs for reducing data needs in load profile analysis. *Appl. Energy* **377**, 124666 (2025).
35. OpenAI. Model optimization. *OpenAI API* https://developers.openai.com/api/docs/guides/model-optimization (2026).
36. Anthropic. Introducing the Model Context Protocol (MCP). https://www.anthropic.com/news/model-context-protocol (2026).
37. Krishnan, N. Advancing Multi-Agent Systems Through Model Context Protocol: Architecture, Implementation, and Applications. Preprint at https://doi.org/10.48550/arXiv.2504.21030 (2025).
38. Hou, X., Zhao, Y., Wang, S. & Wang, H. Model Context Protocol (MCP): Landscape, Security Threats, and Future Research Directions. *ACM Trans. Softw. Eng. Methodol.* 3796519 (2026) doi:10.1145/3796519.
39. Mandal, I. *et al.* Evaluating large language model agents for automation of atomic force microscopy. *Nat. Commun.* **16**, 9104 (2025).
40. EPFL-LBNI. Open-SPM/RePySPM: SPM remote control via Python-Based API Library. https://github.com/Open-SPM/RePySPM (2026).
41. EPFL-LBNI. Open-SPM/Agentic_AFM. https://github.com/Open-SPM/Agentic_AFM (2026).
42. Wang, K., Ruppert, M. G., Manzie, C., Nešić, D. & Yong, Y. K. Adaptive scan for atomic force microscopy based on online optimization: Theory and experiment. *IEEE Trans. Control Syst. Technol.* **28**, 869–883 (2019).
43. Liu, Y. *et al.* Experimental discovery of structure–property relationships in ferroelectric materials via active learning. *Nat. Mach. Intell.* **4**, 341–350 (2022).
44. Liu, Y. *et al.* Autonomous scanning probe microscopy with hypothesis learning: Exploring the physics of domain switching in ferroelectric materials. *Patterns* **4**, 100704 (2023).

45. Krull, A., Hirsch, P., Rother, C., Schiffrin, A. & Krull, C. Artificial-intelligence-driven scanning probe microscopy. *Commun. Phys.* **3**, 54 (2020).

46. Sotres, J., Boyd, H. & Gonzalez-Martinez, J. F. Enabling autonomous scanning probe microscopy imaging of single molecules with deep learning. *Nanoscale* **13**, 9193–9203 (2021).

47. Liu, Y., Ziatdinov, M. A., Vasudevan, R. K. & Kalinin, S. V. Explainability and human intervention in autonomous scanning probe microscopy. *Patterns* **4**, 100858 (2023).

48. Xie, S. & Ren, J. High-speed AFM imaging via iterative learning-based model predictive control. *Mechatronics* **57**, 86–94 (2019).

49. Asmari, N. *et al.* Data-driven control in atomic force microscopy using a genetic algorithm. *Ultramicroscopy* **275**, 114156 (2025).

50. Kammer, C., Nievergelt, A. P., Fantner, G. E. & Karimi, A. Data-Driven Controller Design for Atomic-Force Microscopy. *IFAC-Pap.* **50**, 10437–10442 (2017).

51. Rana, Md. S., Pota, H. R. & Petersen, I. R. The design of model predictive control for an AFM and its impact on piezo nonlinearities. *Eur. J. Control* **20**, 188–198 (2014).

52. Liu, Y. *et al.* Learning the right channel in multimodal imaging: automated experiment in piezoresponse force microscopy. *Npj Comput. Mater.* **9**, 34 (2023).

53. EPFL-LBNI. Open-SPM/OHC. https://github.com/Open-SPM/OHC (2026).

54. Andany, S. H., Nievergelt, A. P., Kangül, M., Ziegler, D. & Fantner, G. E. A high-bandwidth voltage amplifier for driving piezoelectric actuators in high-speed atomic force microscopy. *Rev. Sci. Instrum.* **94**, 093703 (2023).

55. Nievergelt, A. P., Erickson, B. W., Hosseini, N., Adams, J. D. & Fantner, G. E. Studying biological membranes with extended range high-speed atomic force microscopy. *Sci. Rep.* **5**, 11987 (2015).

56. Adams, J. D. *et al.* High-speed imaging upgrade for a standard sample scanning atomic force microscope using small cantilevers. *Rev. Sci. Instrum.* **85**, 093702 (2014).

57. Lowin, J. & Prefect Technologies. FastMCP: The fast, Pythonic way to build MCP servers and clients. GitHub software repository (2026). https://github.com/PrefectHQ/fastmcp (2026).

58. Kubo, S., Umeda, K., Kodera, N. & Takada, S. Removing the parachuting artifact using two-way scanning data in high-speed atomic force microscopy. *Biophys. Physicobiology* **20**, e200006 (2023).

59. Kohl, D., Riel, T., Saathof, R., Steininger, J. & Schitter, G. Auto-tuning PI controller for surface tracking in atomic force microscopy - a practical approach. in 1225 (IEEE, 2016). doi:10.1109/ACC.2016.7526840.

60. Chu, W., Fu, J., Dixson, R., Orji, G. & Vorburger, T. A moving window correlation method to reduce the distortion of scanning probe microscope images. *Rev. Sci. Instrum.* **80**, 073709 (2009).

61. Nečas, D. & Klapetek, P. Gwyddion: an open-source software for SPM data analysis. *Cent. Eur. J. Phys.* **10**, 181–188 (2012).

62. Nečas, D. & Klapetek, P. Gwyddion: an open-source software for SPM data analysis. *Open Phys.* **10**, 181–188 (2012).

**Supplementary Data**

## Agentic AI for operating scientific instruments for nano-scale characterization

Zahra Ayar[1], Marcos Penedo[1], Mahdi Mehdikhani[1], Nahid Hosseini[1], Prabhu Prasad Swain[1], Georg E. Fantner[1]

*Laboratory for Bio- and Nano-Instrumentation, Institute of Bioengineering, School of Engineering, Swiss Federal Institute of Technology Lausanne (EPFL), Lausanne 1015, Switzerland*

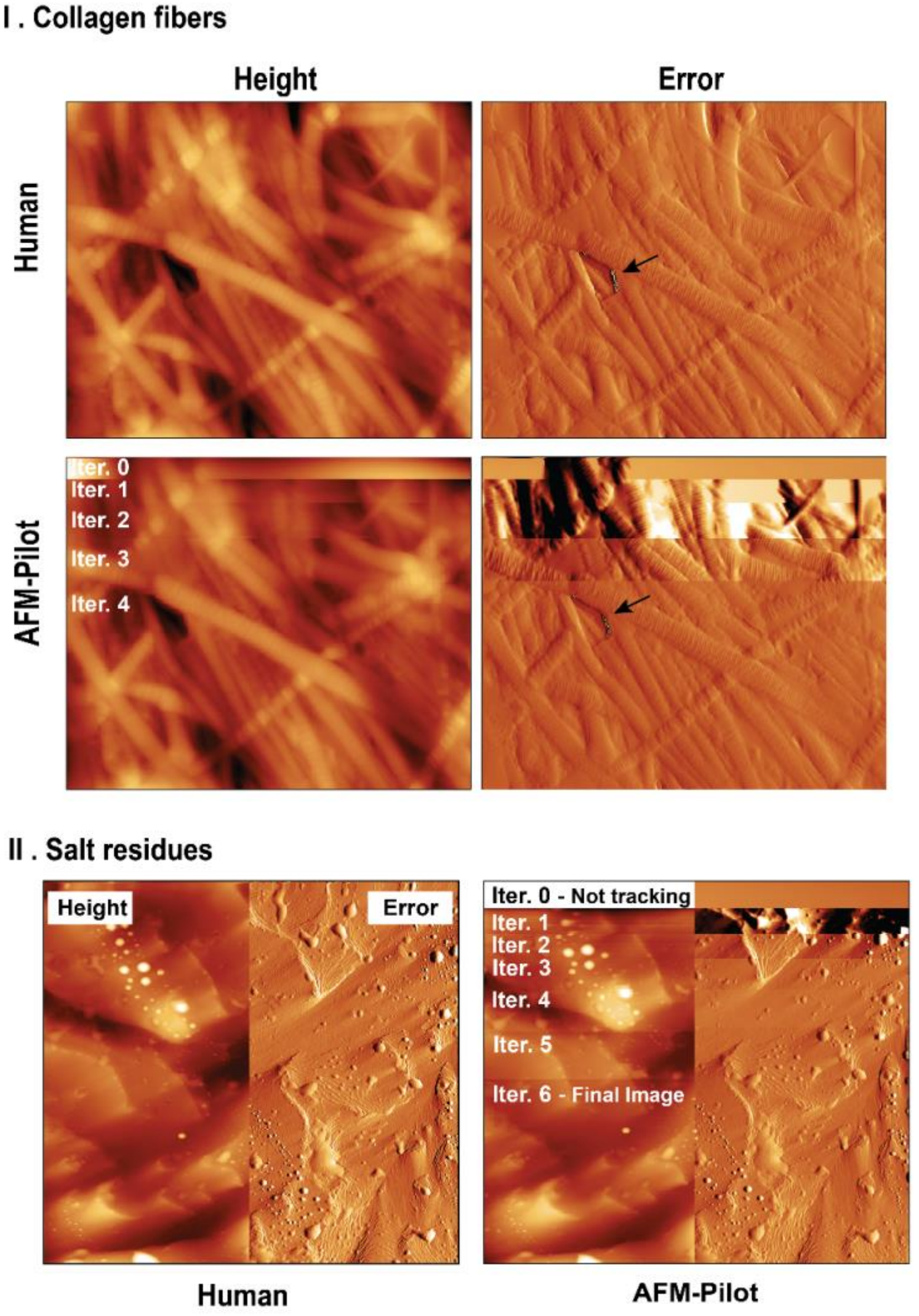

**Figure S1. AFM Pilot tuning across different sample types. I) Collagen-fibril closed-loop tuning. AFM Pilot tunes a collagen-fibril sample from a deliberately detuned start, showing that image-**

**driven parameter optimization extends beyond periodic calibration gratings to a soft biological sample. A local imaging instability remains visible in the final image, indicated by the black arrow, and could not be removed by either the human operator or the AI agent.**

**II) Salt residues on mica after drying PBS. AFM Pilot tunes a heterogeneous salt-residue sample on mica from a detuned starting condition, demonstrating that the same AI-vision-based tuning strategy can be applied to non-periodic and irregular surface features.**

**Table S1.** *Benchmark error percentage across three independent runs per model/framework*

| Model | R1 err%<br>n = 225 | R2 err%<br>n = 200 | R3 err%<br>n = 224 | Mean err% ± SD |
|---|---|---|---|---|
| FT-GPT — text generation (no tools) | 28.0% | 30.5% | 26.8% | 28.4 ± 1.5% |
| FT-GPT-FC — function calling | 23.6% | 22.5% | 23.7% | 23.3 ± 0.7% |
| FT-GPT-FC + tools + AC | 9.3% | 8.5% | 8.9% | 8.9 ± 0.4% |
| Claude Sonnet 4.6 — free-text generation (no tools) | 89.8% | 89.5% | 88.4% | 89.2 ± 0.7% |
| Claude Sonnet 4.6 + tool | 25.3% | 26.0% | 24.1% | 25.1 ± 1.0% |
| Claude Sonnet 4.6 — tools +AC | 4.0% | 4.5% | 5.8% | 4.8 ± 0.9% |
| Claude Opus 4.8 — tools | 6.2% | 8.0% | 4.9% | 6.4 ± 1.6% |
| Claude Opus 4.8 — tools + AC | 1.8% | 2.0% | 3.1% | 2.3 ± 0.7% |
| Claude-MCP + tool + AC (Sonnet 4.6) | 0.0% | 0.0% | 0.0% | 0.0 ± 0.0% |
| Claude-MCP + tool+ AC (Opus 4.8) | 0.0% | 0.0% | 0.0% | 0.0 ± 0.0% |

**Table S2.** *Error taxonomy detail for FT-GPT and FT-GPT-FC.*

| | FT-GPT | | | | FT-GPT-FC | | | |
|---|---|---|---|---|---|---|---|---|
| **Error category** | **R1 (n=225)** | **R2 (n=200)** | **R3 (n=224)** | **Mean % ± SD** | **R1 (n=225)** | **R2 (n=200)** | **R3 (n=224)** | **Mean % ± SD** |
| **Unit / value error** | 35 | 35 | 38 | 52.2% | 39 | 35 | 44 | 78.1% |
| **Ambiguity / missing information** | 8 | 8 | 8 | 13.0% | 12 | 8 | 6 | 17.2% |
| **Extra unintended action** | 19 | 17 | 14 | 27.2% | 1 | 1 | 0 | 1.3% |
| **Wrong tool / module** | 1 | 1 | 0 | 1.1% | 1 | 1 | 3 | 3.3% |
| **Other** | 0 | 0 | 0 | 0.0% | 0 | 0 | 0 | 78.1% |
| **Total failures** | 63 | 61 | 60 | 29.5 ± 1.3% | 53 | 45 | 53 | 23.3 ± 0.7% |

**Table S3.** *Error taxonomy, tool-augmented Claude Sonnet 4.6 model and Opus 4.8.*

| | Sonnet 4.6 model + tools | | | | Opus 4.8 model + tools | | | |
|---|---|---|---|---|---|---|---|---|
| Error category | R1 (n=225) | R2 (n=200) | R3 (n=224) | Mean % ± SD | R1 (n=225) | R2 (n=200) | R3 (n=224) | Mean % ± SD |
| **Unit / value error** | 1 | 1 | 0 | 1.2 ± 0.9% | 0 | 0 | 0 | 0.0 ± 0.0% |
| **Ambiguity / missing information** | 37 | 37 | 41 | 70.7 ± 4.5% | 3 | 3 | 2 | 19.5 ± 1.4% |
| **Extra unintended action** | 2 | 0 | 0 | 1.2 ± 1.7% | 2 | 1 | 1 | 9.9 ± 3.3% |
| **Wrong tool / module** | 17 | 14 | 13 | 26.9 ± 2.3% | 9 | 12 | 8 | 70.7 ± 4.6% |
| **Other** | 0 | 0 | 0 | 0.0 ± 0.0% | 0 | 0 | 0 | 0.0 ± 0.0% |
| **Total failures** | 57 | 52 | 54 | — | 14 | 16 | 11 | — |

**Table S4.** *Error taxonomy, tool-augmented configurations with different model with ambiguity check (AC) layer.*

| Configuration | Correct pass | Ask for clarification | No-tool (true registry gap) | ERROR total | false no-tool | text-only | wrong call | Pass % | Ask % | Error % |
|---|---|---|---|---|---|---|---|---|---|---|
| FT-GPT-FC + AC | 156 | 48 | 0 | 21 | 6 | 15 | 0 | 69.3% | 21.3% | 9.3% |
| FT-GPT-FC + AC | 138 | 45 | 0 | 17 | 3 | 14 | 0 | 69.0% | 22.5% | 8.5% |
| FT-GPT-FC + AC | 147 | 57 | 0 | 20 | 6 | 11 | 3 | 65.6% | 25.4% | 8.9% |
| Claude-MCP + AC (Sonnet 4.6) | 165 | 51 | 9 | 0 | 0 | 0 | 0 | 73.3% | 22.7% | 0.0% |
| Claude-MCP + AC (Sonnet 4.6) | 148 | 46 | 6 | 0 | 0 | 0 | 0 | 74.0% | 23.0% | 0.0% |
| Claude-MCP + AC (Sonnet 4.6) | 156 | 56 | 12 | 0 | 0 | 0 | 0 | 69.6% | 25.0% | 0.0% |
| Claude-MCP + AC (Opus 4.8) | 165 | 51 | 9 | 0 | 0 | 0 | 0 | 73.3% | 22.7% | 0.0% |
| Claude-MCP + AC (Opus 4.8) | 146 | 47 | 7 | 0 | 0 | 0 | 0 | 73.0% | 23.5% | 0.0% |
| Claude-MCP + AC (Opus 4.8) | 157 | 55 | 12 | 0 | 0 | 0 | 0 | 70.1% | 24.6% | 0.0% |

**Information S1. Ambiguity-clarification rule set for the Claude MCP system**

The MCP server controlling the RePySPM interface is initialized with the disambiguation rule set below. These rules were derived from the observed error taxonomy of the fine-tuned and Claude models (Tables S2–S4) and instruct the model to request clarification, rather than commit to a command, in exactly the situations where the bare models fail: SET/GET confusion, module ambiguity for shared parameters, unit uncertainty, and missing values.

- setpoint in Volts (e.g. -0.5 V) -> z_control.set_setpoint
- 'set setpoint' with no value -> ask the user for the value BEFORE calling any tool (SET, not GET)
- when hesitant between sicm and z_control for SET -> choose afm
- 'get / what is / fetch setpoint' with no % context -> z_control.get_setpoint
- update / apply / assign / modify / change / put / configure / define / adjust = SET (use a set_ tool, not get_)
- if a command clearly means SET but no value is given -> ask the user; do NOT call the get_ version while waiting
- NEVER interpret a SET command as a GET just because no value was provided
- 'is approaching / check approaching' -> motors.is_approaching (NOT start_approach)
- 'start / begin / execute approach' -> motors.start_approach
- width / height in micrometres -> pass as metres (multiply by 1e-6)
- hopping mode: scan_parameters for area / pixels / speed; sicm for setpoint / retract_height
- scan commands (X um, NxN, setpoint V, Hz) -> set parameters ONLY; NEVER call scan_up or scan_down during setup

Rules are enforced outside the model, in the MCP server, so they apply identically across model backbones without retraining.

## Information S2. Representative of a session with *AFM Messenger*

Setting resolution to 128

Share

i want resolution 256

Used 2 tools >

Resolution set to 256 × 256 pixels.

i want offset 10 um

Offset X and Y are already set to 10 μm from earlier. Did you want a different value, or should I confirm the current values?

offset in y

Offset Y is already set to 10 μm from earlier. Do you want to change it to something else?

change the scan rate to 4 hz and then set resolution 300 and then start approach

Used 4 tools, loaded tools >

All done:

start approach

Motors__start_approach >

Approach started.

get the resolution

Used 2 tools >

Resolution is 300 × 300 pixels.

set the resolution 256

Used 2 tools >

Resolution set to 256 × 256 pixels.

set resolution 64 and set the width 50 um with 10 um offset in y and then start approach

Resolution set to 256 × 256 pixels.

set resolution 64 and set the width 50 um with 10 um offset in y and then start approach

Used 5 tools >

All done:

- Resolution set to 64 × 64 pixels
- Width set to 50 μm
- Y offset set to 10 μm
- Approach started

set the offset to 20

Did you want the 20 μm offset on X, Y, or both?

Q: Which offset should I set to 20 μm?
A: x

no put 15 degrre rotation

Loaded tools, used a tool >

Rotation set to 15°.

in both

The rotation is a single value for the scan area — there's only one rotation parameter, not separate X/Y rotations. It's already set to 15° for the whole scan frame.

Did you mean something else? For example:

- Apply 15° to another parameter (like offset)?
- Something else I'm missing?

no i want offset for both 10 um

Apr 24

Used 2 tools, loaded tools >

X and Y offsets set to 10 μm.

**Information S3. Representative of a session with *AFM Pilot* and the report generated with system automatically**

## AFM Pilot — Session Report

Closed-loop tuning of LBNI tapping-mode AFM
New tip, free amplitude ≈ 280 mV | Date: 11 June 2026

### Executive Summary

| | |
|---|---|
| **Result** | Good image — best clean balance reached in 10 iterations |
| **Best parameters** | setpoint = 0.22 V, i_gain = 0.015, rate ≈ 1 Hz |
| **Final scores** | parachuting = 0.35 (brief edge touches), ringing = 0.25 (clean) |
| **Free amplitude** | ≈ 280 mV (per operator); final setpoint = 79% of free |
| **Sample** | Patterned steps / pits-pillars (~600 nm features) on a tilted background |
| **Ringing ceiling (mapped)** | i_gain ≈ 0.018–0.022 across setpoints 0.22–0.27; highest clean gain ≈ 0.015 |
| **Stop reason** | Best achievable balance — i_gain at ceiling, setpoint reduction already applied |
| **Advisories** | Creep/tilt NEEDS ACTION (post-processing); hysteresis/double-tip/lost-surface clean |

This tune was performed on a freshly changed tip (free amplitude ≈ 280 mV). Starting from setpoint 0.27 V / i_gain 0.001, *AFM pilot* confirmed tracking, then followed the canonical loop: raise i_gain until ringing, back off i_gain and lower setpoint, re-probe i_gain at the new setpoint. Ten checks converged on the best clean balance the tip/sample allows.

The key result is that the ringing ceiling is set by loop gain (i_gain ≈ 0.016–0.022) and barely moves with setpoint, while lowering setpoint is what reduces parachuting. The chosen operating point — setpoint 0.22 V, i_gain 0.015 — sits just under the ringing ceiling at the highest setpoint that still tracks the steep feature edges, with only brief 2–4 px edge parachuting remaining.

### Iteration-by-iteration trace

| It | Setpt (V) | i_gain | Para AI/np | Ring AI/np | Action | Note |
|---|---|---|---|---|---|---|
| **1** | 0.27 | 0.001 | 0.55/0.60 | 0.00/0.00 | i_gain ×5 → 0.005 | Tracking confirmed: clear step features, Height fwd/bwd track, parachuting at descending edge (rail -0.43V). Climb i_gain. |
| **2** | 0.27 | 0.005 | 0.40/0.60 | 0.15/0.08 | i_gain ×3 → 0.015 | Parachuting easing (0.55→0.40). Ringing edge-recovery only. Keep climbing. |
| **3** | 0.27 | 0.015 | 0.45/0.60 | 0.20/0.36 | i_gain → 0.022 (probe) | Still clean ringing by AI; flats clean. Probe toward ceiling. |
| **4** | 0.27 | 0.022 | 0.30/0.00 | 0.65/0.59 | RINGS — i_gain↓ + sp↓ | CEILING HIT. Frame-wide oscillation filling flat regions, all lines. Back off i_gain AND lower setpoint. |
| **5** | 0.25 | 0.014 | 0.40/0.59 | 0.25/0.30 | re-probe i_gain → 0.018 | Lower setpoint cleared the ringing. Re-probe gain upward at new setpoint. |
| **6** | 0.25 | 0.018 | 0.45/0.01 | 0.35/0.45 | RINGS — back to 0.015 | Ceiling at 0.25 also ~0.015–0.016. Ringing in flat regions, line-correlated. Settle just under. |
| **7** | 0.25 | 0.015 | 0.45/0.59 | 0.20/0.34 | lower setpoint → 0.22 | Clean again at 0.015. Parachuting persists → spend setpoint for force. |
| **8** | 0.22 | 0.015 | 0.35/0.58 | 0.00/0.29 | re-probe i_gain → 0.018 | More force helped: parachuting 0.45→0.35, ringing CLEAN 0.00. Re-probe gain. |
| **9** | 0.22 | 0.018 | 0.35/0.01 | 0.35/0.40 | RINGS — back to 0.015 | Ceiling ~0.015–0.016 at this setpoint too. Line-correlated oscillation in flats. Settle. |
| **10** | 0.22 | 0.015 | 0.35/0.33 | 0.25/0.34 | STOP — best balance | 199-line frame. Parachuting brief 2-4px edge touches; |

| It | Setpt (V) | i_gain | Para AI/np | Ring AI/np | Action | Note |
|---|---|---|---|---|---|---|
| | | | | | | ringing clean, flats clean; features well resolved. |

*Legend:* Para/Ring shown as AI score / numpy score. AI drives decisions; numpy is a parallel guide only. *Green row = final operating point.*

### Advisory: artifacts beyond parachuting / ringing

Mandatory advisory section. *AFM Pilot* checks creep, hysteresis, double-tip and lost-surface on every frame; the tuning loop acts only on parachuting and ringing, so the rest are reported here for the operator.

| Creep / sample tilt | NEEDS ACTION |
|---|---|
| **AI detection** | DETECTED most iterations (severity 0.30–0.45) |
| **What AI saw** | Both Height [Fwd] and [Bwd] show a broad background ramp (~300–500 nm across the frame, up to ~1300 nm early on) equally in both scan directions — sample tilt or slow piezo creep, not a tip artifact. Diminished over the session as drift settled. |
| **Recommended action** | Apply background plane subtraction / tilt correction in post-processing (Gwyddion plane-fit or built-in flatten). Optionally allow more thermal settling before a final scan. |
| **Impact on tune** | Cosmetic only — does not affect tuning. Correct before exporting/quantifying the image. |

| Hysteresis (piezo) | CLEAN |
|---|---|
| **AI detection** | NOT detected at any iteration (AI 0.00 throughout) |
| **What AI saw** | numpy hysteresis fired sporadically (mean_shift 4–10 px on some frames) but AI never confirmed. Consistent with the known numpy false-positive on patterned/periodic samples. |
| **Recommended action** | No action. Trust AI over numpy on hysteresis for this sample type. |
| **Impact on tune** | None. |

| Double tip | CLEAN |
|---|---|
| **AI detection** | NOT detected at any iteration (AI 0.00 throughout) |
| **What AI saw** | No ghost/shadow copies of features; pit/pillar edges single and sharp in both fwd and bwd across all frames. |
| **Recommended action** | No action. Clean single-tip imaging confirmed. |
| **Impact on tune** | None. |

| Lost surface | CLEAN |
|---|---|
| **AI detection** | NOT encountered during the tune |
| **What AI saw** | Once engaged at setpoint 0.27, the tip tracked throughout; no featureless ramps or sustained full-line rail saturation beyond the parachuting edge events. (Note: earlier in the day on a different tip, free-air/engagement issues did occur — not part of this successful run.) |
| **Recommended action** | No action. If lost-surface recurs, lower setpoint gently in small steps until features return — do not retract/re-approach blindly. |
| **Impact on tune** | None. |
| **CLEAN** | Not detected by AI; no action needed. |
| **NUMPY ONLY** | numpy fired but AI did not confirm; likely false-positive. |
| **NEEDS ACTION** | AI flagged it; operator should act (here: post-processing). |

## Findings

| Observation | Detail |
|---|---|
| **Ringing ceiling is gain-limited, ~constant across setpoint** | At setpoints 0.27, 0.25 and 0.22 the onset of frame-wide ringing was all near i_gain 0.016–0.022. The cantilever/loop sets a hard gain ceiling that lowering setpoint did not raise much. |
| **Lower setpoint helped parachuting, not the gain ceiling** | Dropping 0.27→0.22 cut parachuting 0.55→0.35 and even cleared ringing at fixed gain — more force improves tracking, but the gain ceiling stayed put. |

| Observation | Detail |
|---|---|
| **AI now catches ringing correctly** | After the prompt rewrite, the AI flagged the frame-wide ringing at iter 4 (0.65) and the milder line-correlated ringing at iters 6 and 9 (0.35) — matching the operator's eye and numpy's high-band trend, instead of under-calling it as before. |
| **Numpy parachuting noisy, AI steady** | numpy parachuting swung 0.00–0.60 frame-to-frame (per-row buffer sensitivity), while AI stayed a steady 0.35–0.55. AI vision drove the decisions; numpy used only as a guide. |
| **Residual parachuting is brief edge-only** | At the stop point, parachuting is 2–4 px rail touches at steep feature edges — about the floor for this tip/sample without pushing into ringing. |

### Loop procedure

1. **Confirm tracking.**
   Iter 1: real step features in Height, fwd/bwd track the feature → tracking confirmed before any scoring.
2. **Raise i_gain until it rings.**
   Iters 1–4: climbed 0.001 → 0.022; ringing appeared at 0.022 (frame-wide). That is the ceiling for setpoint 0.27.
3. **When it rings: drop i_gain a bit AND drop setpoint a bit.**
   Iter 4→5: i_gain 0.022→0.014 and setpoint 0.27→0.25. Ringing cleared.
4. **Re-probe i_gain at the new setpoint.**
   Iters 5–9: probed up at 0.25 and again at 0.22; ceiling ~0.015–0.016 each time. Lower setpoint reduced parachuting.
5. **Stop at best clean balance.**
   Iter 10: setpoint 0.22, i_gain 0.015 — highest clean gain, highest setpoint that tracks, ringing clean, only brief edge parachuting left.

### Best parameters

**Setpoint = 0.22 V** (≈ 79% of 280 mV free amplitude)
**i_gain = 0.015** (just below the ringing ceiling ≈ 0.016)
**Scan rate ≈ 1 Hz**

### Open items for next session

- Apply plane-fit / tilt correction to the saved image — the ~300–500 nm background ramp (creep/tilt) is the only flagged advisory.
- detector_config.json still carries stale setpoint values from a previous cantilever (comfort floor 0.5 V, working 1.52 V). For this 280 mV tip the floor should be ~0.168 V. The AI's repeated "setpoint below comfort floor" note is harmless but stale — worth updating.

- numpy parachuting per-row aggregation swings widely (0.00–0.60) on essentially the same image. Low priority; AI vision is authoritative, but a steadier numpy guide would reduce disagreement re-checks.
- If brief edge parachuting (0.35) matters for this dataset, a slightly lower setpoint (~0.20 V) could be

Information S4. A section of AFM-Pilot report after imaging collagen fibril where it consider the collagen banding (with confirmation of operator) in feedback ringing.

## 2. Iteration Table

| # | Setpoint (V) | i_gain | Parach. AI/np | Ring AI/np | Action / Note |
|---|---|---|---|---|---|
| — | 1.90 | 0.001 | — | — | Start (operator-set manually): free amp ~2.1 V, setpoint 90%. AI: NOT TRACKING (free-air, blank). |
| — | 1.75 | 0.001 | 0.00 / 0.45 | 0.00 | Still not tracking — smooth ~700 nm bow, Error railed. Setpoint down. |
| — | 1.65 | 0.001 | 0.55 / 0.01 | 0.15 | TRACKING — large collagen fibril with real banding resolved. |
| 1 | 1.65 | 0.001→0.003 | 0.55 / — | 0.15 | i_gain ×3. AI: wiggles track features (banding), not frame-wide ringing. |
| 2 | 1.65 | 0.003→0.009 | 0.45 / 0.00 | 0.15 | i_gain ×3. AI: 'periodic banding is genuine structure, not ringing artifact'. |
| 3 | 1.65 | 0.009 | 0.00 | 0.00 | Scanned onto bare region between fibrils — NOT TRACKING. Setpoint down. |
| — | 1.60 | 0.009 | 0.15 | 0.45 / 0.06 | Re-engaged. AI 'ringing' 0.45 — OPERATOR: good image, slight parachuting. |
| 4 | 1.60 | 0.009→0.011 | 0.00 | 0.45 / 0.00 | i_gain ↑ to trim slight parachuting → parachuting clean. |

| ★ | **1.60** | **0.011** | **0.00** | **banding (eye)** | **OPERATOR: good image, parachuting clean, rings sharp. AI 'ringing' 0.45 = collagen banding.** |
|---|---|---|---|---|---|

Status of every advisory artifact this run, per the mandatory reporting rule — listed even when clean.

## 3. Advisory Section

**Feedback ringing (PRIMARY FINDING — recurring)**

- AI scored ringing 0.45 frame-wide at i_gain 0.009 and 0.011; numpy scored 0.00–0.06 throughout.
- Operator confirmed on the live screen: the periodic Error signal is the collagen's real D-band banding, not feedback ringing. The image is good.
- The AI named the banding correctly ('periodic banding is genuine structure, not ringing artifact') but still scored that same periodic signal as ringing and prescribed cutting i_gain. The sample-context note is not enough to stop the mis-score.
- Action taken: did NOT cut i_gain to chase this 'ringing'. Operator eye treated as authority. This is the second consecutive run with the same mis-call — detector fix is the top open item.

**Parachuting**

- Mild parachuting on the steep descending fibril flanks at i_gain 0.009 (operator noted 'slight parachuting'). Raising i_gain to 0.011 cleared it (AI and numpy both 0.00 after). No setpoint reduction needed for it.

**Hysteresis (numpy + AI)**

- Numpy flagged up to 1.00 on the broad fibril hump (fwd/bwd shift of the large smooth mound); AI never confirmed. Likely numpy false-positive, consistent with operator notes.

**Creep**

- Not detected by AI in any iteration. No action.

**Double tip**

- Not detected across all iterations — clean single-tip imaging confirmed.

**Lost surface**

- Encountered three times (setpoint 1.90 and 1.75 at start; bare region between fibrils after iter 2). Each handled in software by gently lowering setpoint until real collagen features reappeared — never retracted. Tracking restored each time.

**Information S5. Representative of a session with *AF M Doctor***

Identified corrupted images and flagged problematic interpolation artifacts for review

All 12 diagnosed and processed — PNGs saved next to each `.gwy` in `2026-07-07_exp00`. Here's the rundown, with scar-correction counts as a health check:

| File | Crop | Scars / px | Notes |
|---|---|---|---|
| 0000 | 90 rows | 0 | clean |
| 0001 | — | 19 / 2037 | ⚠ high; a 10-row cluster (128–137) |
| 0002 | — | 2 / 54 | clean |
| 0003 | — | 5 / 282 | clean |
| 0004 | — | 0 | clean |
| 0005 | — | 0 | clean |
| 0006 | — | 6 / 623 | fine |
| 0007 | — | 11 / 1809 | ⚠ moderately high |
| 0008 | — | 10 / 1659 | ⚠ + AI flagged **striping** |
| 0009 | 74 rows | 4 / 240 | clean |
| 0010 | — | **47 / 5650** | ⚠ ⚠ ~40-row corrupted band at bottom |
| 0011 | — | **29 / 3316** | ⚠ ⚠ ~38-row corrupted band at bottom |

Most came out clean. Three things I want to flag honestly rather than declare it all done:

**0010 and 0011** — the AI and the scar counts agree there's a solid *band* of corrupted scan lines along the bottom (~40 rows each). `remove_scars` "fixed" them by interpolating, but interpolating 40 contiguous rows means that strip is essentially synthesized, not real data. These are better **cropped** like a dead band. The edge detector missed them because they're bright/noisy rather than flat. If you upload `0010`, I'll pick the right `crop_bottom` and re-run both.

**0008** — the AI specifically flagged periodic striping, which is exactly what the new `destripe` step is for. Once you restart to load it, I'll re-run 0008 with `destripe` and compare.

**0001 / 0007** — higher scar counts than the clean ones; worth an eyeball for any over-correction.

Want to upload `0010`, `0008`, and `0001` so I can verify those three and dial in the fixes? The rest look solid from the numbers.

**Videos S1.** Live AFM Pilot session, showing natural-language command interpretation, ambiguity clarification, and iterative closed-loop tuning to the final image.